\documentclass[letterpaper, 10 pt, conference]{ieeeconf}  % Comment this line out if you need a4paper

\IEEEoverridecommandlockouts                              % This command is only needed if 
\usepackage{graphics} % for pdf, bitmapped graphics files
\usepackage{epsfig} % for postscript graphics files
\usepackage{times} % assumes new font selection scheme installed
\usepackage{amsmath} % assumes amsmath package installed
\usepackage{amssymb}  % assumes amsmath package installed

\usepackage{amsthm}
\usepackage{amsmath}
\usepackage[ruled,vlined]{algorithm2e}
\usepackage{subfig}
\usepackage{multirow}
\usepackage{dirtytalk}
\usepackage{xcolor}
\usepackage{url}

\usepackage{algpseudocode}% http://ctan.org/pkg/algorithmicx

\title{\LARGE \bf
Domain Concretization from Examples: \\
Addressing Missing Domain Knowledge via Robust Planning
}

\author{Akshay Sharma, Piyush Rajesh Medikeri and Yu Zhang% <-this % stops a space
\thanks{Akshay Sharma, Piyush Rajesh Medikeri and Yu Zhang are with the School of Computing, Informatics
and Decision Systems Engineering, Arizona State University, Tempe, AZ. {\{\tt\small ashar204, pmediker, yu.Zhang.442\} @asu.edu}}

}

\begin{document}

\maketitle
\thispagestyle{empty}
\pagestyle{empty}

%%%%%%%%%%%%%%%%%%%%%%%%%%%%%%%%%%%%%%%%%%%%%%%%%%%%%%%%%%%%%%%%%%%%%%%%%%%%%%%%
\begin{abstract}

The assumption of complete domain knowledge is not warranted for robot planning and decision-making in the real world. It could be due to design flaws or arise from domain ramifications or qualifications. In such cases, existing planning and learning algorithms could produce highly undesirable behaviors. This problem is more challenging than partial observability in the sense that the agent is unaware of certain knowledge, in contrast to it being partially observable: the difference between known unknowns and unknown unknowns. In this work, we formulate it as the problem of \textit{domain concretization}, an inverse problem to domain abstraction. Based on an incomplete domain model provided by the designer and teacher traces from human users, our algorithm searches for a candidate model set under a minimalistic model assumption. It then generates a robust plan with the maximum probability of success under the set of candidate models. In addition to a standard search formulation in the model-space, we propose a sample-based search method and also an online version of it to improve search time. We tested our approach on IPC domains and a simulated robotics domain where incompleteness was introduced by removing domain features from the complete model. Results show that our planning algorithm increases the plan success rate without impacting the cost much.

% Most planning and learning systems assume the complete knowledge about the state space. 

% However, this might not always be true in situations where the agent model does not have the knowledge about the missing information or some state features that it can't perceive. In such cases planning and learning algorithms could produce behaviors that are highly undesirable. In this work we formulate this problem of "Inconceivability" and address this issue by considering the scenarios where the robot model is missing some state features. Using the optimal action traces from human expert, our algorithm finds out if model is missing some state features and if so, our algorithm generates a set of models that has the possibility to be the correct model. Using this set of models, our algorithm then finds a robust plan that is valid for the maximum number of models among the set of possible correct models. We have tested our algorithm on various IPC domains and also on a simulated robotics domain.

\end{abstract}

%%%%%%%%%%%%%%%%%%%%%%%%%%%%%%%%%%%%%%%%%%%%%%%%%%%%%%%%%%%%%%%%%%%%%%%%%%%%%%%%
\section{INTRODUCTION}

Most planning agents rely on complete knowledge of the domain, which could be catastrophic when certain knowledge is missing in the domain model. The fact that the agent is unaware of some domain knowledge makes this problem different and more challenging than partial observability where the agent knows what is missing. That is the difference between known unknowns and unknown unknowns. For example, missing certain state features means that the agent would perceive all states as if those features do not exist. In such cases, traditional planning algorithms can create the same plan under very different scenarios. For similar reasons, standard RL~\cite{RLRS}, IRL~\cite{IRL}, \cite{MaxentIRL}, and intention recognition algorithms \cite{IRMao}, \cite{IRSchrempf} could be easily misled as a result of incomplete domain knowledge.

We refer to the process required to address this problem as \textit{\textbf{domain concretization}}, which is inverse to domain abstraction. In this paper, we focus on state abstraction and leave action abstraction in future work. Even though state abstractions in planning have significant computational advantages, if not engineered properly, they can lead to unsound and incomplete domain specifications~\cite{angelic, sidMeta}. Incomplete domain specification also arises naturally from domain ramifications \cite{FingerRamification} and qualifications \cite{QualificationsMc}, \cite{GinsbergQual}. %which may be viewed as intentional state abstractions. 

%There could be multiple reasons for a model to miss some state features. One of that could be intentional or unintentional introduction of state abstraction. 

%Even though state abstraction has significant computational advantages, if not designed correctly, it can result in some necessary features becoming inconceivable. Another reason could be due to partial observability of the human mental state. Since the robot can’t perceive the human mental state directly, in many cases the state feature representing it is inconceivable. In some situations, a broken or a missing sensor could also lead to the corresponding state feature being inconceivable.

Consider a manufacturing domain where a robot is tasked to deliver a model to a human worker, which is produced by a 3D printer. Depending on how long the model has cooled down before the delivery, the robot is expected to either apply a coolant first or deliver it immediately to the human worker. 
However, if the temperature is an unknown domain feature to the robot, it may result in immediate delivery regardless of the temperature, which could lead to safety risks.

In this paper, we restrict the domain concretization problem to deterministic domains. We formulate this problem using a STRIPS-like language \cite{STRIPS} and provide search methods for candidate models. Our search algorithm generates candidate models based on the incomplete model provided by the domain designer and teacher traces from human users.
This problem is first formulated as a search problem in the model space.
To expedite the search, we then present a sample-based search method by considering only models that are consistent with the traces. Additionally, we present an online version that is more practical and has a computational advantage as it uses only one trace in each iteration and maintains a small set of models for future refinements. 

For planning, our algorithm generates a robust plan that achieves the goal under the maximum number of candidate models. This is done by transforming the planning problem into a Conformant Probabilistic Planning problem \cite{PFF}.
% In cases where this problem leads to huge risks, our solution will increase the robustness of learning and planning systems by considering Domain Concretization for generating safer plans, even though such plans may be more costly in some cases.  
We tested our algorithm on various IPC domains and a simulated robotics domain where incompleteness was introduced by removing specific predicates from the complete domain model. Results show that the robust plan generated by our algorithm increases the plan success rate under the complete model without impacting the cost much.

\section{RELATED WORK}

The ongoing research on state abstraction has established its necessity and computational advantages. There exists work where authors developed abstractions that retained properties of the ground domain. The authors in \cite{abelNear} have studied abstractions that produce optimal behaviors similar to those with the ground domain. In \cite{sidMeta}, the authors have investigated and categorized several abstraction mechanisms that retain the properties of the ground domain like the Markov property. The fact that the agent's domain model is missing some domain features in our work makes the model an abstraction of the ground model. Domain concretization is the inverse of domain abstraction: instead of searching for an abstraction, here we are trying to recover the ground domain model from the abstract (incomplete) model.

% In our problem this type of abstraction is not helpful and instead can produce highly undesired results.
% This problem is harder than partial observability in the sense that the agent is not even aware of the existence of certain knowledge, in contrast to it being just partially observable; the difference between known unknowns and unknown unknowns.

It may be attempting to solve the domain concretization problem using a POMDP formulation \cite{POMDP}, or reinforcement learning methods with POMDPs \cite{RLPOMDP}, where the state is partially observable and the agent uses observations to update its belief about the state of the world. Note however that a POMDP still requires complete knowledge about the ground state space, which is not available in our problem formulation due to missing domain knowledge. More specifically, this means that, neither the space of belief states nor the observation function is known, unlike that in POMDP. Hence, POMDP solutions cannot be used to solve our problem.

% There exists work where the authors have analyzed the domain flaws to find the correct domain model. In \cite{blindSpot} the authors have considered domain flaws due to mismatches between simulation and the real world. These mismatches could be due to some unintentional and incorrect abstractions of the real-world in the simulated environment which causes some blind spots in the real world. Their algorithm predicts these blind spots in a Reinforcement Learning environment. The algorithm uses oracle feedback like demonstration and corrections to learn a model that predicts these blind spots. In our work, in addition to the detection of missing information, our algorithm can generate a robust plan which is executable irrespective of the missing information.

The agent's model in our problem could be considered an approximate domain model and hence planning in such a case becomes a type of model-lite planning \cite{modelliteRao}. Many existing approaches have considered planning in such approximate (or incomplete) domain models. Authors in \cite{Default}, \cite{Tuan} introduce planning systems that can generate plans for an incomplete domain where actions could be missing some preconditions or effects. Our problem can be viewed as a more general problem where the information of possible missing knowledge (e.g., possible preconditions) is missing. % from the domain.

The idea of learning action model using plan traces has been there for quite some time. While some authors have considered refining incomplete domain models \cite{RefiningIP}, \cite{ModelLiteCaseBased}, others have focused more on learning action models from scratch \cite{AMExMaxSAT}, \cite{LearningComplex}. Authors in \cite{AMATL} have used transfer learning to learn the action model using a small amount of training data. In \cite{AMAN}, authors have gone one step further by developing an algorithm that can learn action models even when the traces are noisy. One common assumption in all these methods is the complete knowledge of the state space, about which, the agent in our problem, does not have.

\section{DOMAIN CONCRETIZATION}

\subsection{Problem Analysis}
Consider the motivating example of the manufacturing domain mentioned in the previous section. The robot is expected to deliver a 3D model to a human, produced by a 3D printer. The human teacher performs some demonstrations to teach the robot how the task is supposed to be done. The human has the complete domain model $M^*$ but the robot's domain model is incomplete $\widetilde{M}$, where the temperature is an unknown feature. In such a case the traces (demonstrations) are generated by the teacher using $M^*$ and {\em projected} onto $\widetilde{M}$ that is observed by the robot. In the projection, the robot does not (or cannot) observe the missing feature since it is not present in its domain model (i.e., the robot would  consider it irrelevant to the task scenario and/or  lacks the appropriate sensor for feature extraction). Before further discussion let us first list the assumptions:

\begin{enumerate}
    \item The human teacher is rational and the traces are optimal in the complete domain model $M^*$.
    
    \item The complete domain model $M^*$ is deterministic.
    
    \item  The robot still knows about all the possible actions.
    
    \item The unknown features are {\it completely} missing from the domain model.
    
    \item The unknown features do not appear in the goal, since the goal is often provided by the human user who is also the teacher.
\end{enumerate}

Given the assumptions mentioned above, an observed (projected) trace would still contain all the actions in the teacher's trace but the state trajectory associated with it will be viewed differently.  %that in projected traces the start state could be missing the unknown feature. 
To make this clearer, let us have a look at an example trace. Consider the following teacher trace/plan that is observed by the robot in our motivating example:

\begin{itemize}
    \item Plan $z_1$: there is a {\it hot} 3D model (that is just printed) in the start state. The action sequence is: $\langle$ {\tt apply\_coolant}, {\tt deliver} $\rangle$.
    
    % \item In the second trace $z_2$, there is a hot 3D model in the start state. The corresponding action sequence is- $\langle$ {\tt apply\_coolant2}, {\tt deliver} $\rangle$. The action {\tt apply\_coolant1} is different from the action {\tt apply\_coolant2} but have the same cost.
    
    % \item In the third trace $z_3$, the start state has a 3D model that is not hot. Here, the action sequence is - $\langle$ {\tt deliver} $\rangle$. 
\end{itemize}

In the teacher's perspective,
the state sequence for this action sequence is:
$\langle$ \{{\tt{hot}}\}, \{{\tt{wet}}\}, \{{\tt{wet}, \tt{delivered}}\} $\rangle$,
assuming that {\tt{apply\_coolant}} has two effects, one being making the model {\tt{wet}}, and the other being removing {\tt{hot}}.

The robot observes the action and state sequences above except for the {\tt{hot}} feature, 
which requires the {\tt apply\_coolant} action to be first applied in the complete model. 
Since the feature is completely missing in the robot's model, the {\tt apply\_coolant} action would also be missing it (i.e., the robot is unaware that this action would remove {\tt hot} from the state or, in other words, make the model cooler).
Hence, the state sequence that is observed by the robot would be:
$\langle$ \{\}, \{{\tt{wet}}\}, \{{\tt{wet}, \tt{delivered}}\} $\rangle$.
Hence, it would consider that the {\tt apply\_coolant} action is unnecessary,
assuming that the action {\tt deliver} does not require the model to be {\tt wet}.

Hence, given the observed initial state \{\}, in the robot's model $\widetilde{M}$, the optimal plan for this ``same'' scenario is (note again that the robot does not observe the {\tt hot} feature in the initial state):

\begin{itemize}
    \item Plan $z_0$: there is a 3D model in the start state. The action sequence is: $\langle$ {\tt deliver} $\rangle$.
    
    % \item In the second trace $z_2$, there is a hot 3D model in the start state. The corresponding action sequence is- $\langle$ {\tt apply\_coolant2}, {\tt deliver} $\rangle$. The action {\tt apply\_coolant1} is different from the action {\tt apply\_coolant2} but have the same cost.
    
    % \item In the third trace $z_3$, the start state has a 3D model that is not hot. Here, the action sequence is - $\langle$ {\tt deliver} $\rangle$. 
\end{itemize}

Obviously, there is an inconsistency here:
with the assumption that the teacher is rational and 
cost captured by plan length, 
the teacher is expected to never choose anything longer than $z_0$. 
% such a ino
% In the projection of these traces $\widetilde{z_1}$, $\widetilde{z_2}$ and $\widetilde{z_3}$ respectively, the robot will perceive the start state as the same for all the traces, that is the 3D model regardless of the temperature. The action sequences for $\widetilde{z_1}$, $\widetilde{z_2}$, and $\widetilde{z_3}$ are the same as $z_1$, $z_2$ and $z_3$, respectively, except that the state trajectories will be different with the missing feature not present in the robot's view. 
The solution to the problem of {\it domain concretization} is hence hinged on addressing this type of inconsistency, which we refer to as {\em plan inconsistency}.
Let us analyze this further to understand how trace inconsistency can help us concretize our domain models and its limitations. 
Consider a teacher's plan $z_1$ (under the complete model $M^*$).
Denote the robot's optimal plan (under the incomplete model $\widetilde{M}$) for the same observed/projected initial ($\widetilde{s_0}$) and goal states ($g$) as $z_0$.
Two possibilities are here:
%The observed (projected) trace is also $z$.

%\begin{enumerate}
    %\item %There are no missing features in the initial state (and it is assumed no missing features in the goal state). 
    %Two possibilities here:
    % The start state of the trace $z$ is the same as the start state of the trace $z'$. In the manufacturing domain example consider $z$ = $z_1$ and $z'$ = $z_2$. In this case, the projected traces will also have the same start states as in $\widetilde{z_1}$ and $\widetilde{z_2}$. 
   % There are two possibilities in this case.
    \begin{enumerate}
        \item Cost($z_1$) $\neq$ Cost($z_0$) --
        Inconsistency: the teacher has chosen a more or less costly plan than the optimal plan in the robot's model, which should not occur if the robot's model is complete. In fact, given our assumptions above, the only possibility here is that Cost($z_1$) $>$ Cost($z_0$). And that is because when a feature is completely missing in the domain model, it will be missing as both preconditions and effects. Hence, the missing features would only make the incomplete model less constrained to satisfy the goal. %Model concretization must be performed here.
        % \begin{enumerate}
        
        % this is not possible since the 
        %  If the complete domain $M$ is deterministic, then this is not possible. This is because of the assumption that the traces are optimal in the complete domain model $M$. If that is the case there can not be two optimal traces with different costs for the same start state and goal.
        
        % \item
        % If the complete domain $M$ is stochastic, then this is possible. In that case, we can not make a definite answer about whether the traces are inconsistent. This is because for stochastic domain, different optimal cost could either be due to the stochastic nature or could be due to the unknown predicate missing.
        % \end{enumerate}
        
        \item Cost($z_1$) $=$ Cost($z_0$) -- Here, there are two possibilities:
        
        \begin{enumerate}
        \item $z_1$ is a valid plan for $(\widetilde{s_0}, g)$ in $\widetilde{M}$: Undetermined (deemed consistent): $z_1$ could be a valid plan in an incomplete model $\widetilde{M}$ or it could be that $\widetilde{M}$ is complete. Since we cannot decide which case is true here, we will have to wait until inconsistencies are detected. 
        %Each of the traces $\widetilde{z}$ and $\widetilde{z}'$ are optimal (executable and achieves goal with minimum cost) in the incomplete domain $\widetilde{M}$ for same start state. In that case, the model $\widetilde{M}$ could still be incomplete, but there is no way to find out. In such cases, the traces can not be called inconsistent.
        
        \item $z_1$ is not a valid plan for $(\widetilde{s_0}, g)$ in $\widetilde{M}$ -- Inconsistency: $z_1$ is not a valid plan in the robot's model but a valid plan in the complete model.  
        
        % Out of all the traces, there exists one that is not optimal in $\widetilde{M}$. In this case, the trace become \textit{inconsistent} with the robot's model $\widetilde{M}$. In the example of the manufacturing domain, consider trace $z_1$ and its projections $\widetilde{z_1}$. In this case, the action sequence contains an action {\tt apply\_coolant1}. This action generates a precondition required for the execution of the action {\tt deliver}. It cools down the temperature of the 3D model which is a precondition for the execution of {\tt deliver}. Since in $\widetilde{M}$ temperature of the 3D model is an unknown feature, this precondition does not exist. Hence, the optimal plan in $\widetilde{M}$ would just be $\langle$ {\tt deliver} $\rangle$. Now, the trace is no longer optimal and this makes the trace inconsistent as according to the assumption traces should be optimal. 
        \end{enumerate}
        
    \end{enumerate}
    
    In our problem solution, we address only the cases when any plan inconsistencies above are detected. 
    However, even when all the plan inconsistencies are addressed, it does not mean that the complete model is recovered due to the case of 2.(a) above.
    On the other hand, as more teacher traces are provided, our solution is expected detect more inconsistencies as long as the complete model is not recovered.

\subsection{Background}

Given our focus on deterministic domains, we use a STRIPS-like language to define our problem. Here, a planning problem is defined by a triplet $P$ $=$ $\langle s_0$, $g$, $M\rangle$, where $s_0$ is the start state, $g$ is the set of goal prepositions that must be \textbf{T} (true) in the goal state and $M$ is the domain model. The domain model $M$ $=$ $\langle O$, $R \rangle$, where $R$ is the set of predicates and $O$ is the set of operators. The set of prepositions $F$ and the set of actions $A$ are generated by instantiating all the predicates in $R$ and all the operators in $O$ respectively. Hence, we can also define $M$ $=$ $\langle A$, $F \rangle$. A state is either the set of prepositions $s$ $\subseteq$ $F$ that are true or $s_\bot$ $=$ \{$\bot$\}. The state $s_\bot$ is a dead state and once it is reached, the goal can never be achieved. The actions change the current state by adding or deleting some prepositions. Each action a $\in$ $A$ is specified by a set of preconditions $Pre(a)$, a set of add effects $Add(a)$ and a set of delete effects $Del(a)$, where $Pre(a)$, $Add(a)$ and $Del(a) \subseteq F$. For a model $M$, the resulting state after executing plan $\pi$ in state $s$ is determined by the transition function $\gamma$, which is defined as follows:
\begin{equation}
\gamma (\pi, s)=
\begin{cases}
  s & \text{if }
       \!\begin{aligned}[t]
       \pi = \langle\rangle; \\
       \end{aligned}
\\
  \gamma(\langle a\rangle, \gamma(\pi', s))  & \!\begin{aligned}[t]
       \pi = \pi' \circ \langle a \rangle. \\
       \end{aligned}
\end{cases}
\end{equation}

In our problem, we use the \textit{Generous Execution (GE)} semantics as defined in \cite{Tuan} where if an action $a$ is not executable, it does not change the world state $s$. Hence, the transition function $\gamma$ for an action sequence with a single action $a$ and state $s$ under GE semantics is defined as follows:
\begin{equation}
\gamma (\langle a \rangle, s)=
\begin{cases}
  ( s \setminus Del(a)) \cup Add(a) & \text{if }
       \!\begin{aligned}[t]
       Pre(a) \subseteq s; \\
       \end{aligned}
\\
  s & \text{otherwise.}
\end{cases}
\end{equation}

A plan $\pi$ is a valid plan for a problem $P$ $=$ $\langle s_0$, $g$, $M\rangle$ iff $\gamma (\pi, s_0) \supseteq g$. The cost of a plan $\pi$ is the cumulative cost of all the actions in $\pi$. To make our discussion simpler, we assume that all the actions incur one unit cost. In such a case, the cost of a plan will simply be equal to the plan length. One may also note that our method will still work even if the cost for each action is different.

\subsection{Motivating Example}

\begin{figure}%
    \centering
    \noindent\fbox{%
    \parbox{0.434\textwidth}{%
    \tt{
    \footnotesize{
    
    (:action open\_box\\ 
    :parameters (?b $–$ box)\\
    :precondition (and (handempty))\\
    :effect (and (box\_open ?b)))\\\\
    (:action grasp\\
     :parameters (?i $–$ item)\\
     :precondition (and (on\_shelf ?i) (handempty))\\           
    %  \hspace*{1cm}(handempty))\\
     :effect (and (holding ?i) (not (on\_shelf ?i))\\
            %   \hspace*{1cm}(not (on\_shelf ?i))\\
             \hspace*{1cm}(not (handempty))))\\\\
    (:action place\\
      :parameters (?i1 - item ?b $–$ box)\\
      :precondition (and (holding ?i1) (box\_open ?b)\\ 
    %   \hspace*{1cm}(box\_open ?b)\\
              \hspace*{1cm}(box\_empty ?b))\\
      :effect (and (item\_packed ?i1) (handempty)\\ 
    %   \hspace*{1cm}(handempty)\\
	        \hspace*{1cm}(on\_top ?i1 ?b) (not (holding ?i1))\\
	       % \hspace*{1cm}(not (holding ?i1))\\
            \hspace*{1.cm}(not (box\_empty ?b))))\\\\
    (:action stack\\
       :parameters (?i1 - item ?i2 - item ?b - box)\\
       :precondition (and (holding ?i1) (box\_open ?b)\\
    %   \\\hspace*{1cm}(and (holding ?i1)\\ 
                    \hspace*{1cm}{\textbf{(not\_fragile ?i2)}} (on\_top ?i2 ?b))\\
                %   \hspace*{1cm}(box\_open ?b) \\
                %   \hspace*{1cm}(on\_top ?i2 ?b)) \\
       :effect (and (item\_packed ?i1) (handempty)\\
    %   \\\hspace*{1cm}(and (item\_packed ?i1) \\
    %   \hspace*{1cm}(handempty)\\
	\hspace*{1cm}(on\_top ?i1 ?b) (not (on\_top ?i2 ?b))\\
% 	\hspace*{1cm}(not (on\_top ?i2 ?b))\\
           \hspace*{1cm}(not (holding ?i1))))

        }
        }
    }%
    }
    % \subfloat[\centering Actual Domain ]{{\includegraphics[width=4cm]{img0.png} }}%
    % \qquad
%         \noindent\fbox{%
%     \parbox{0.3\textwidth}{%
%     \tt{
%     \footnotesize{
%         (:action open\_box\\ 
%     :parameters (?b $–$ box)\\
%     :precondition (and (handempty))\\
%     :effect (and (box\_open ?b)))\\
%     (:action grasp\\
%      :parameters (?i $–$ item)\\
%      :precondition (and (on\_shelf ?i)\\            (handempty))\\
%      :effect (and (holding ?i)\\
%               (not (on\_shelf ?i))\\
%               (not (handempty))))\\
%     (:action place\\
%       :parameters (?i1 - item ?b $–$ box)\\
%       :precondition (and (holding ?i1)\\ (box\_open ?b)\\
%               (box\_empty ?b))\\
%       :effect (and (item\_packed ?i1)\\ (handempty)\\
% 	  (on\_top ?i1 ?b) (not (holding ?i1))\\
%              (not (box\_empty ?b))))\\
%     (:action stack\\
%       :parameters (?i1 - item ?i2 - item\\ ?b - box)\\
%       :precondition (and (holding ?i1)\\ 
%                               (box\_open ?b) (on\_top ?i2 ?b))
%       :effect (and (item\_packed ?i1) (handempty)
% 	(on\_top ?i1 ?b) (not (on\_top ?i2 ?b))
%             (not (holding ?i1))))
%         }
%         }
%     }%
%     }
    % \subfloat[\centering Incomplete Domain ]{{\includegraphics[width=4cm]{img1.png} }}%
    % \qquad
    \captionsetup[subfigure]{labelformat=empty}
    \subfloat[\centering %Simulation View
    ]{{\hskip3pt\includegraphics[width=7.96cm]{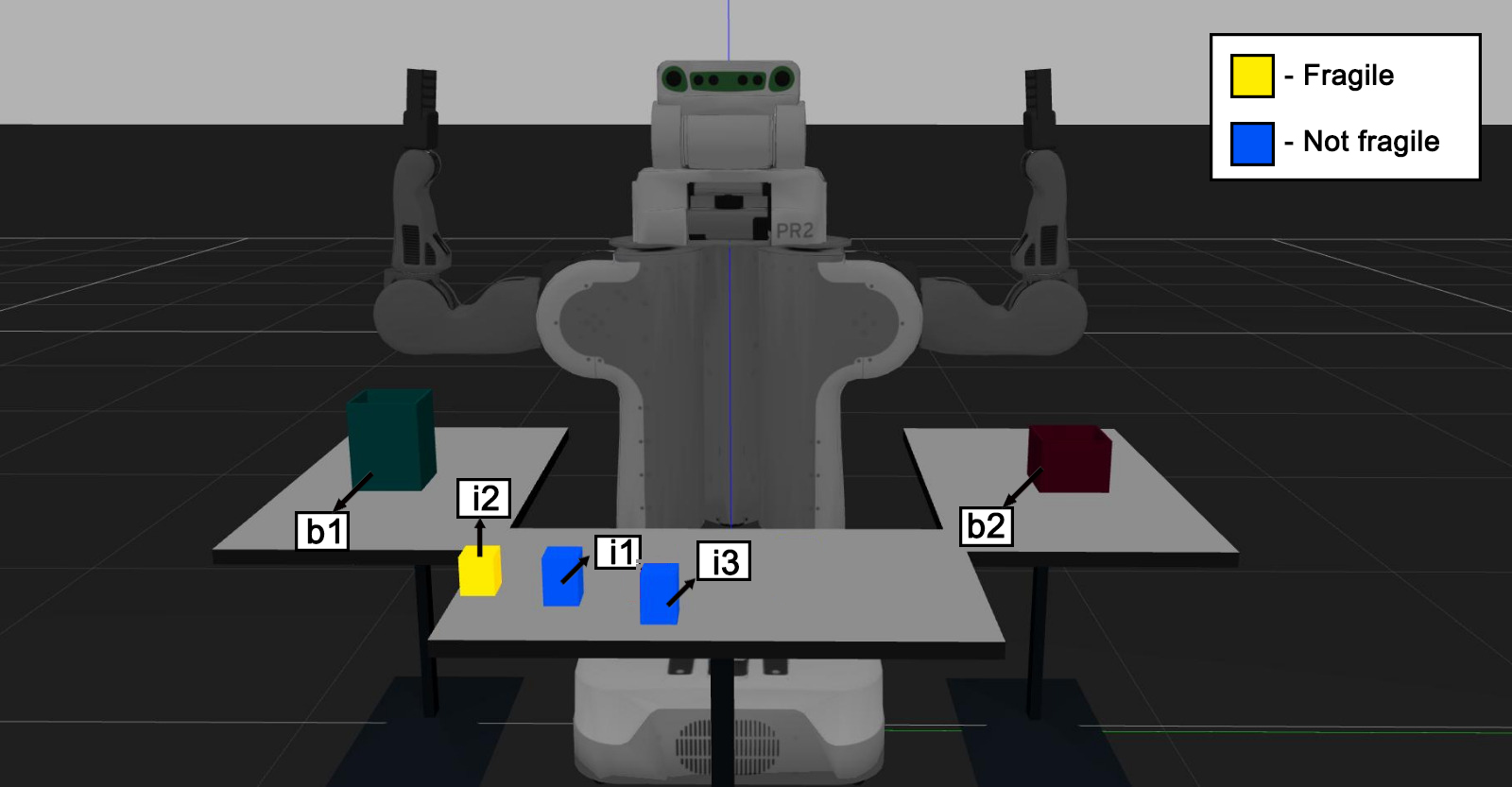} }}%
    \vskip-10pt
    \caption{Packing domain as a runnig example}%
    \label{fig:domain}%
    \vskip-10pt
\end{figure}

\textit{Complete domain model} (denoted by $M^*$): The specification of a packing domain is shown in Fig. \ref{fig:domain} to motivate our problem. The domain has $4$ operators. As the name suggests, {\tt{open}}\_{\tt{box}} is used to open boxes where items are to be placed. The {\tt{grasp}} operator is used to grasp an item. {\tt{place}} and {\tt{stack}} are used to put items into boxes. {\tt{place}} is used when the box is empty and {\tt{stack}} is used when the box already contains some items. The goal is to store all items in boxes using the minimum number of boxes. 
Some of the items may be fragile.
For these items, they must not be stacked on. 
%the robot must ensure that it shouldn't put any other item above the fragile item. 
To avoid such situations, % there is a predicate, 
a predicate {\tt{not}}\_{\tt{fragile}} is introduced foritems that are not fragile.
%whether the item is not fragile and is a precondition for stack action. 

\textit{Incomplete domain model} (denoted by $\widetilde{M}$): In the robot's domain model that is incomplete, the incompleteness may be due to missing the predicate {\tt{not}}\_{\tt{fragile}} (shown in bold in Fig. \ref{fig:domain}). In such a case, the robot may choose a plan in which 
it stacks an item over a fragile item. %might stack some item over a fragile which is highly undesirable.

\subsection{Problem Formulation}

The incomplete model 
%(the robot's model in Fig. \ref{fig:domain}) 
 $\widetilde{M}$ $=$ $\langle \widetilde{O}$, $\widetilde{R} \rangle$ is incomplete in the sense that it is missing some predicates present in the complete domain $M^* = \langle O^*$, $R^* \rangle$. 
%In our packing domain, the robot's model is missing the predicate of {\tt{not}}\_{\tt{fragile}}.
% In our coffee delivery example mentioned before the robot's model is incomplete because it is missing the predicate representing the hotness (temperature) of the coffee from the actual model. 
% If $\widehat{R}$ is the set of missing predicates, then $\widetilde{R}$ $\subseteq$ $R^*$ such that $\widetilde{R} = R^* \setminus \widehat{R}$, $\widetilde{Pre(o)} = Pre(o) \setminus \widehat{R}$, $\widetilde{Add(o)} = Add(o) \setminus \widehat{R}$ and $\widetilde{Del(o)} = Del(o) \setminus \widehat{R}$.
Denote $\widehat{R}$ as the set of missing predicates, such that:
\begin{itemize}
\item
$\widetilde{R}$ $\subseteq$ $R^*$ and $\widetilde{R} = R^* \setminus \widehat{R}$
\item
$\widetilde{Pre(o)} = Pre(o) \setminus \widehat{R}$
\item
$\widetilde{Add(o)} = Add(o) \setminus \widehat{R}$
\item
$\widetilde{Del(o)} = Del(o) \setminus \widehat{R}$
\end{itemize}

% If $\widehat{R}$ is the set of missing predicates,
% \begin{itemize}
% \item
% $\widetilde{R}$ $\subseteq$ $R^*$ such that $\widetilde{R} = R^* \setminus \widehat{R}$
% \item
% $\widetilde{Pre(o)} = Pre(o) \setminus \widehat{R}$
% \item
% $\widetilde{Add(o)} = Add(o) \setminus \widehat{R}$
% \item
% $\widetilde{Del(o)} = Del(o) \setminus \widehat{R}$
% \end{itemize}

\newtheorem{definition}{Definition}
\begin{definition}
  The problem setting of \textbf{Domain Concretization} is defined as a setting where the agent has only access to an incomplete domain model $\widetilde{M}$ and teacher traces under $M^*$. %which is missing a set of predicates $\widehat{R}$ from the complete domain model $M^*$.
\end{definition}

% The problem of {\it Domain Concretization} is defined as a scenario where the robot has incomplete model $\widetilde{M}$ and the human's model is $M^*$, the true complete model. 
We assume that the human user knows $M^*$ so 
%Since the human knows $M^*$, 
he/she can provide a set of successful teacher traces $\zeta^*$ based on $M^*$. Each trace $z^*$ $\in$ $\zeta^*$ is a tuple $\langle s_0^*, g, \tau \rangle$ where $g$ and $s_0^*$ are the goal and initial state respectively and $\tau$ is an action sequence $\langle a_1, a_2, ...  a_n \rangle$ where each $a_i$ $\in$ $A^*$. Since the robot has incomplete predicate set $\widetilde{R}$, it observes {\it incomplete traces} $\widetilde{\zeta}$ = $\langle \widetilde{s_0}$, $g$, $\tau \rangle$,  where $\widetilde{s_0} = s_0^* \setminus \widehat{R}$. We assume that the robot  still has knowledge about all the actions, and hence $\tau$ would not be affected by the incompleteness. Additionally, we also assume that the missing predicate cannot be present in the goal state since the goal is provided by the human user. 
%Hence, $g$ will also not be affected by incompleteness.

% The missing predicate can't be present in the goal because in that case there will be multiple contradicting states which will be impossible to achieve by any plan at the same time. For instance, if the goal of the robot is to deliver hot coffee and the preposition representing temperate of coffee (hot/cold) is missing, it is impossible for a robot to generate a plan that delivers hot and cold coffee ate the same time. Hence, $g$ will also not be affected by incompleteness.

\begin{definition}
  %Given $\widetilde{M}$ and $\widetilde{\zeta}$, 
  The problem of Planning under Incomplete Domain Knowledge (PIDK) is defined as $\widetilde{P}$ $=$ $\langle \widetilde{s_0}, g, \widetilde{M}, \widetilde{\zeta} \rangle$, which is the problem of generating a plan that has the highest probability of success for $P^*$ $=$ $\langle s_0^*, g, M^* \rangle$.
\end{definition}

\subsection{Candidate Model Generation}

We search for the candidate models that have the minimum number of new predicates and the minimum number of changes introduced into the given incomplete model $\widetilde{M}$. One of the motivations for this assumption could be attributed to the principle of \textit{Occam's Razor} \cite{Occam2}. Another reason is due to the fact that here, the model-space search problem has infinite possible solutions as we can always introduce more dummy predicates and still make the model generate the given traces. This is somewhat similar to the unidentifiability problem in reward learning \cite{IRL1}, \cite{SSingh}, \cite{occam} and we call it the problem of \textit{model unidentifiability}.

% We transform the problem of generating candidate models as a search problem in the model-space. Each state in the search space is a domain model ($M$) generated by adding a set of new predicates to different positions in the incomplete domain model ($\widetilde{M}$). We define a variable $\sigma$ whose value indicates the number of new predicates that are going to be added. This parameter is set to 1 in the beginning and is gradually increased if no models are found within a predefined threshold. Further, for all new predicates, we define a set $X$ which is the set of possible typed predicates generated (from all possible type combinations). These are the predicates that we add to the domain model $\widetilde{M}$. Since these predicates could also be present in the start state, we consider multiple possible start states where each could be defined as $s_0 = \widetilde{s_0} \cup \mu$ where $\mu \in {\mathfrak P} (U)$; $U$ is the set of prepositions instantiated using all new predicates that have been added to $M$ and ${\mathfrak P} (U)$ represents power-set of $U$. Each candidate model $M$ would be verified against all these possible start states. The model $M$ is accepted as a candidate model if it passes the goal test for at-least one $s_0$. Given all this, the model-space search problem is defined as follows:

We transform the problem of generating candidate models to a search problem in the model space. We define a variable $\sigma$ whose value indicates the number of new predicates that are going to be added. Initially, $\sigma$ is set to 1 and is gradually increased if no models are found for the current value. Further, we define a set $X$ as the set of possible typed predicates (generated from all possible type combinations) for each new predicate. Each state in the search space is a domain model ($M$) generated by adding one or more predicates from $X$ to one or more possible missing positions in the incomplete domain model ($\widetilde{M}$). Since these predicates could also be present in the start state, each candidate model $M$ is verified against one or multiple possible start states $s_0$, based on the prepositions added to $\widetilde{s_0}$ (details in the search section). The model $M$ is accepted as a candidate model if it passes the model test below for at-least one $s_0$. Our model-space search is defined as follows:
\begin{itemize}
\item
\textbf{Initial State:} $\widetilde{M}$

\item
\textbf{Action Set ($\Lambda$):} $\alpha^{Pre(o)}_{\chi}$ $($or  $\alpha^{Add(o)}_{\chi} / \alpha^{Del(o)}_{\chi}) \in \Lambda$,

\begin{center}
$\forall \chi \in X$ and $\forall o \in O$ where $O \in M$.    
\end{center}
 
The actions in the model-space search represent a predicate $\chi$ being added to the $Pre(o)$ $($or $Add(o) / Del(o))$ of the current model $M$.

%is the action of the search problem which represents predicate $\chi$ being added to $Pre(o)$ (or $Add(o) / Del(o)$ respectively) to generate the next model $M'$. 

\item
\textbf{Successor Function ($T$):} $T(M, \alpha^{Pre(o)}_{\chi}) = M'$. 

$T$ produces new model $M'$ where $R' = R \cup \chi$ and $Pre(o') = Pre(o) \cup \chi$ where $o' \in O'$ and $o \in O$. Similarly, we can define $T(M, \alpha^{Add(o)}_{\chi})$ and $T(M, \alpha^{Del(o)}_{\chi})$.

\item

\textbf{Model (Goal) Test:} $C_1(M)$ $\land$ $C_2(M)$ $\land$ $C_3(M)$.

$C_1(M)$, $C_2(M)$ and $C_3(M)$ are defined as follows-
\begin{itemize}
\item 
\textit{Plan Validity Test}, $C_1(M)$ is True if : 
\begin{equation}
 \gamma^M(\tau, s_0) \supseteq g; 
 \forall \langle \widetilde{s_0}, g, \tau \rangle \in \widetilde{\zeta}
\end{equation}
 This ensures that the traces are executable and achieve the goal under $M$. This condition essentially detects the type of inconsistency mentioned in case 2.(b) of the problem analysis section (III.A).

\item
\textit{Well-Justification Test}, $C_2(M)$ is True if : 
\begin{equation}
 \forall a_i \in \tau, \gamma^M(\tau \setminus \{a_i\}, s_0) \nsupseteq g;  
 \forall \langle \widetilde{s_0}, g, \tau \rangle \in \widetilde{\zeta}
\end{equation}

This ensures that the traces are well-justified \cite{Fink1992FormalizingPJ} in $M$, which means that if any action is removed from the trace the goal will not be achieved. This condition detects the type of inconsistency mentioned in case 1 of the problem analysis section (III.A).

% This is a direct consequence of our assumption that the traces are optimal.

% \item
% $C_2(M)$ is True if : $\forall \langle \widetilde{s_0}$, $g$, $\tau \rangle \in \widetilde{\zeta}$, 
% \begin{equation}
%  \forall a_i \exists p | p \in Add(a_i) \implies \exists a_j | p \in Pre(a_j), 
% \end{equation}
% where $a_i, a_j \in \tau$, $1 \leq i \leq n$ and $i+1 \leq j \leq n (n = |\tau|)$.
% This condition ensures that each action generates at least one preposition that is required as a precondition of some subsequent action. Since we assume that traces are optimal, each action has to be useful in the given trace.

\item 
\textit{Plan Optimality Test}, $C_3(M)$ is True if : 
\begin{equation}
Cost(\pi^M) = Cost(\tau); 
\forall \langle \widetilde{s_0}, g, \tau \rangle \in \widetilde{\zeta}
\end{equation}
where $\pi^M$ is the optimal plan for problem $P$ $=$ $\langle s_0$, $g$, $M\rangle$; $Cost(\pi^M)$ and $Cost(\tau)$ represents plan cost of $\pi^M$ and $\tau$ respectively. This condition ensures that the traces are optimal under $M$. This condition detects the type of inconsistency mentioned in case 1 of the problem analysis section (III.A).

\end{itemize}
\end{itemize}

In the packing domain, let $\tau$ = $\langle${\tt open\_box} {\tt b1}, {\tt grasp} {\tt i1}, {\tt place} {\tt i1} {\tt b1}, {\tt grasp} {\tt i2}, {\tt stack} {\tt i2} {\tt i1} {\tt b1}, {\tt open\_box} {\tt b2}, {\tt grasp} {\tt i3}, {\tt place} {\tt i3} {\tt b2}$\rangle$. Applying the algorithm to $\tau$ and $\widetilde{M}$, $C_3$ will fail because, optimal plan, $\pi^M$ = $\langle${\tt open\_box} {\tt b1}, {\tt grasp} {\tt i1}, {\tt place} {\tt i1} {\tt b1}, {\tt grasp} {\tt i2}, {\tt stack} {\tt i2} {\tt i1} {\tt b1}, {\tt grasp} {\tt i3}, {\tt stack} {\tt i3} {\tt i2} {\tt b1}$\rangle$, is less costly than $\tau$. This indicates that the trace is inconsistent as mentioned in the case 1 of the problem analysis (III.A) section.

\textbf{Theorem 1:} $C_1(M)$ and $C_3(M)$ are necessary and sufficient to ensure that the model $M$ can generate $\tau$, $\forall \langle \widetilde{s_0}, g, \tau \rangle$ $\in$ $\widetilde{\zeta}$.

% If and only if a model $M$ satisfies conditions $C_1(M)$ and $C_3(M)$ for traces $\widetilde{\zeta}$, model $M$ can generate $\tau$ for $\widetilde{s_0}$ and $g$, $\forall \langle \widetilde{s_0}$, $g$, $\tau \rangle$ $\in$ $\widetilde{\zeta}$.

\textbf{Proof:} If $C_1(M)$ is true then $\gamma^M (\tau, s_0) \supseteq g$ which means $\tau$ is a valid plan for $\langle s_0$, $g$, $M \rangle$. Also, if $C_3(M)$ is true then cost of optimal plan of $M$ equals cost of $\tau$ which implies $\tau$ has minimum cost under $M$. Hence, by definition $\tau$ is the optimal plan for the given problem under model $M$. Since this is true for all $\tau$, it can be concluded that $M$ can generate $\tau$ $\forall \langle \widetilde{s_0}$, $g$, $\tau \rangle$ $\in$ $\widetilde{\zeta}$. Similarly, if a model M can generate $\tau$ for a problem $\langle s_0$, $g$, $M \rangle$ then $\tau$ an optimal plan for the problem. Hence, by definition of optimal plan, $\tau$ is valid and has minimum cost which implies that it satisfies $C_1(M)$ and $C_3(M)$. This proves that the model test is sound and complete.

% Since we assume traces $\widetilde{\zeta}$ to be optimal in the actual model we can say that $M$ can generate $\tau$ $\forall \langle \widetilde{s_0}$, $g$, $\tau \rangle$ $\in$ $\widetilde{\zeta}$. 

% Similarly, if a model M can generate $\tau$ for a problem $\langle s_0$, $g$, $M \rangle$ then $\tau$ an optimal plan for the problem. Hence, by definition of optimal plan, $\tau$ is valid and has minimum cost. Since $\tau$ is valid, it satisfies $C_1(M)$. Also, $\tau$ is an optimal plan which implies $Cost(\tau)$ = $Cost(\pi^M)$ where $\pi^M$ is any optimal plan under model $M$. Hence $\tau$ satisfies $C_3(M)$. Hence, every model $M$ that can generate $\tau$ satisfies both $C_1(M)$ and $C_3(M)$. This proves that our goal test is sound and complete.

\textbf{Theorem 2:} $C_2(M)$, is a necessary condition for a trace $\tau$ where $ \langle \widetilde{s_0}, g, \tau \rangle$ $\in$ $\widetilde{\zeta}$ to be optimal.

\textbf{Proof:} This could be proven by contradiction. Assume that model $M$ doesn't satisfy $C_2(M)$ but still $\tau$ is an optimal plan for $M$. This implies that for some action $a_i$, $\tau \setminus \{a_i\}$ is a valid plan. Hence, there is a valid plan $\tau \setminus \{a_i\}$ which has lesser cost than $\tau$ as $Cost(\tau \setminus \{a_i\}) = Cost(\tau) - Cost(a_i)$. This implies that $\tau$ is not an optimal plan which contradicts the initial assumption. Hence proved.

% Generating optimal plan for a problem is a hard problem in computer science. Since $C_3(M)$ involves generation of optimal plan for model $M$, the algorithm would become very slow if it checks for optimal plan for every model. To improve this our algorithm checks for $C_3(M)$ only if $C_2(M)$ fails. This reduces the search time by a huge amount. Also, using the result from Theorem 2 we can say that this will not change the set of candidate models generated. 

$C_3(M)$ requires the computation of an optimal plan which is costly. Hence, we check for $C_3(M)$ only if $C_2(M)$ fails to reduce the search time. % by a significant amount.
Theorem 2 ensures that this process will not lose any candidate models.

\textbf{Search:} This formulation can be solved by any standard search algorithm and we chose a uniform cost search. The cost of the path from $M$ to $M'$ is the number of changes introduced into the domain model $M$ to generate $M'$. The search starts with $\widetilde{M}$ as the initial state for the model-space search. Using the action set $\Lambda$ and the transition function $T$, the set of next models is generated and put into a priority queue. The model with the least cost is then popped and passed through the model test. The model is tested with multiple $s_0$ for each trace. If $X'$ ($X' \subset X$) is the set of predicates added to the current domain model $M$, we generate $U$ as a set of all the prepositions that can be instantiated from each predicate in $X'$. Now, $s_0 = \widetilde{s_0} \cup \mu$ where $\mu \in {\mathfrak P} (U)$. Each $\mu$ is chosen such that $|\mu|$ is minimum. First we test with $|\mu|$ = 0 and if the model test is passed we do not go further. If it fails, we test with every $\mu$ such that $|\mu|$ = 1. This process goes on until a predefined threshold is reached and in that case model test fails. Setting this threshold to 0 denotes an assumption that missing predicate could never be present in the start state. 

Now, If the model test is passed, all the models having the same cost are popped and tested. Each model that passes the model test becomes a candidate model. If the model test fails, the set of next models is generated in a similar way as mentioned before. This process continues until some model passes the model test or the cost increases to be above a predefined threshold. In the latter case, $\sigma$ is increased by $1$ and the whole search starts again. In the end, we obtain sets of models such that, within each set of models, $s_0$ used to satisfy the model test remains the same. Each model could be in multiple sets based on the $s_0$'s that satisfy the model test. The candidate model set $\mathcal{M}$ is a weighted union of these sets of models based on how many times each model appears in those sets.

% The algorithm keeps on adding predicates to the model until the goal test is passed. If for any model goal test is passed, the algorithm checks for all the models with same depth and return all the models that satisfies the goal test with that depth. There is a threshold limit to the number of changes that is set in the beginning. If none of the model satisfy the goal test within the threshold limit, we increase $\sigma$ and start whole search again. At the end we get sets of models that satisfy the goal test. Each set of models has different prepositions ($\mu$) added to their start states.

In our example packing domain in Fig \ref{fig:domain}, we start with $\widetilde{M}$ as the model $M$ and $\sigma$ = 1. Denote the missing predicate as {\tt pred\_1}. Since $M$ fails $C_3(M)$, the search is started by generating $X$. $X$ includes all the possible typed predicates like {\tt(pred\_1} {\tt?b} - {\tt box)}, {\tt(pred\_1} {\tt?b} - {\tt box} {\tt?i} - {\tt item)}, etc. Using $X$, we generate the set $\Lambda$ that includes all the possibilities where the predicates in $X$ can be added, like $\alpha^{Pre(open\_box)}_{\chi}$ $\forall \chi \in X$, which means $\chi$ will be added to the preconditions of operator {\tt open\_box}. Using the current model $M$, $T$ and $\alpha \in \Lambda$, we generate model $M'$. %Next, $M'$ is model tested and based on the result the search continues as explained above.

% Computational complexity:

% **We might be able to reduce problem of computing plan robustness for incomplete domain to our problem and then prove it is \#P-Complete**.

\subsection{Generating candidate models using sample-based search}

% One obvious way to perform the model search is by brute-forcing through all the possible models, which is computationally expensive and not scalable. 
% % as it considers all the possibilities where the given predicate could be added,
% Hence, we present a sample-based search which reduces the search space by a considerable amount and makes the search faster.

% Based on the conditions below, a set of actions $\Lambda'$ is returned to guide the search algorithm and generate a smaller set of candidate models for further steps. The action set $\Lambda'$ now include sets of actions that represent multiple simultaneous changes. The procedure also decides what $s_0$ to use for a model $M$ in the model test. The search process is similar as before except that if a model fails the model test, instead of returning false, it returns the set of actions $\Lambda'$. Instead of checking the model $M$ for all possible $s_0$'s, we check for the ones that are returned along with the actions. 

% Since these changes are necessary to make the model pass the model test,  changes that would not lead to a candidate model are ignored. The returned action set is as follows:

To contain the computational complexity, we further present a sample-based search to reduce the search space. 
The idea is to return information for refining the model to satisfy the traces only, instead of checking all possibilities.  
The action set $\Lambda'$ now is a set of actions that each encodes multiple simultaneous changes. Such a process should also decide what $s_0$ to use for a model $M$ in the model test. This process is similar as before except that if a model fails the model test, instead of returning false, it returns a set of actions, $B \subseteq \Lambda'$ that will be used to generate models for the next step. 
Instead of checking the model $M$ for all possible $s_0$'s, we check for the ones that are returned along with the action set.
The returned action set $B$ is as follows:

% Hence, we present a heuristic-based approach to search through the model space. Here instead of considering all the possible actions in the action set, the algorithm chooses a smaller set of actions that could possibly lead to the goal. This reduces the search space by a considerable amount and makes the search faster.

% For each $M$ that does not satisfy goal conditions, the heuristics provide the set of actions $\Lambda'$ which guides the search algorithm and generates the next set of model space states. The action set $\Lambda'$ now also includes pairs of actions which implies multiple simultaneous changes. The heuristics also tells what $s_0$ to use for a model $M$ for the goal test. The search process is similar as before except that if a model-space state fails the goal test then instead of returning false, it returns the set of possible actions $\Lambda'$. Since these changes are necessary to make the model pass the goal conditions, it would help the algorithm to ignore the changes that won't lead to the goal. The returned action set is as follows:

\begin{itemize}
\item
\textit{Unsatisfied Precondition}: Here, the trace is not executable in $M$ because of some unsatisfied precondition. 
This means, $C_1(M)$ is false and, for an action $a_i \in \tau$, $1 \leq i \leq|\tau|$, $Pre(a_i) \nsubseteq \gamma^M(\langle a_1, a_2, ... a_{i-1}\rangle, s_0)$. Then, the returned action set $B = \{\{\alpha^{Add(o_j)}_{\chi}\}\}$, $\forall a_j \in \tau$ and $\forall \chi \in X$, where $o_j$ is the operator corresponding to action $a_j$ and $1 \leq j \leq i-1$. Also, possible additions to start state $s_0$ will be $Pre(a_i) \setminus \gamma^M(\langle a_1, a_2, ... a_{i-1}\rangle, s_0)$. This is because missing preposition could either be present in the add effect of any of the previous actions or in the start state. This addresses the type of trace inconsistency mentioned in case 2.(b) of the problem analysis section (III.A).

\item
\textit{Un-justified action}: This happens when some action in the trace is not well-justified in $M$.
This means, $C_2(M)$ is false and for some action $a_i$, $\gamma^M(\tau \setminus \{a_i\}, s_0) \supseteq g$. Then, the returned action set $B = \{\{\alpha^{Add(o_i)}_{\chi}, \alpha^{Pre(o_j)}_{\chi}\}\} \cup \{\{\alpha^{Add(o_i)}_{\chi}, \alpha^{Pre(o_j)}_{\chi}, \alpha^{Del(o_j)}_{\chi}\}\}$ $\forall a_j \in \tau$ and $\forall \chi \in X$. $o_i$ and $o_j$ are operators corresponding to actions $a_i$ and $a_j$ respectively and $i+1 \leq j \leq |\tau|$. Intuitively, this generates $B$ such that $a_i$ cannot be removed from $\tau$ which makes it well-justified under $M$. This addresses the type of trace inconsistency mentioned in case 1 of the problem analysis section (III.A).
% {\color{blue} The set in $\Lambda'$ represents that these changes will be introduced simultaneously in the model $M$ to obtain $M'$.}

\item
\textit{Sub-optimal Trace}: This happens when the optimal plan under $M$ is shorter than the trace.
Here, $C_3(M)$ becomes false, which means there has to be some action that was not possible in $\tau$ under $M^*$ but was possible in $\pi^M$ under $M$. Hence, the operator corresponding to that action is missing some precondition. In such a case, $\exists a_i, a'_i$ such that $a_i \in \tau$ and $a'_i \in \pi^M$ and $a_i \ne a'_i$, $1 \leq i \leq n$ $(n = |\pi^M|)$. Then, the returned action set $B = \{\{\alpha^{Pre(o_j)}_{\chi}\}\} \cup \{\{ \alpha^{Pre(o_j)}_{\chi}, \alpha^{Del(o_j)}_{\chi} \}\}$ $\forall a'_j \in \pi^M$, where $o_j$ is the operator corresponding to $a'_j$ and $i \leq j \leq n$. This addresses the type of trace inconsistency mentioned in case 1 of problem analysis section (III.A).

% Here, $C_3(M)$ becomes false, which means there has to be some action that wasn't possible in $\tau$ but was possible in $\pi^M$. Hence, the operator corresponding to that action is missing some precondition. In such a case, $\exists a_i, a'_i$ such that $a_i \in \tau$ and $a'_i \in \pi^M$ and $a_i \ne a'_i$, $1 \leq i \leq n$ $(n = |\pi^M|)$. Then, the possible action set $\Lambda' = \{\alpha^{Pre(o_j)}_{\chi}\}$ $\forall a'_j \in \pi^M$ where $o_j$ is operator corresponding to $a'_j$ and $i \leq j \leq n$. For each $\{\alpha^{Pre(o_j)}_{\chi}\} \in \Lambda'$ a tuple $\{\langle \alpha^{Pre(o_j)}_{\chi}, \alpha^{Del(o_j)}_{\chi} \rangle\}$ is also added to $\Lambda'$.

\end{itemize}

For the packing domain in Fig. \ref{fig:domain}, consider $\tau$ = $\langle${\tt open\_box b1}, {\tt grasp i1}, {\tt place i1 b1}$\rangle$. If some model $M$ is missing the predicate {\tt(box\_open ?b)}, $C_2(M)$ will fail because the goal will be achieved even after the deletion of {\tt open\_box b1} from $\tau$. In that case, $B = \{\{\alpha^{Add(open\_box)}_{\chi}, \alpha^{Pre(grasp)}_{\chi}\}$,$ \{\alpha^{Add(open\_box)}_{\chi}$, $ \alpha^{Pre(place)}_{\chi}\}\} \cup \{\{\alpha^{Add(open\_box)}_{\chi}, \alpha^{Pre(grasp)}_{\chi}, \alpha^{Del(grasp)}_{\chi}\}$, $ \{\alpha^{Add(open\_box)}_{\chi}, \alpha^{Pre(place)}_{\chi}, \alpha^{Del(place)}_{\chi}\}\}$  $\forall \chi \in X$.

% Using the actions in $\Lambda'$ for each model-space state we generate the successive states in the search and search continues as mentioned in the previous section.

% Using the actions in $B$ for each model-space state the successive states are generated in the search and it continues as mentioned in the previous section.

\textbf{Theorem 3:} (\textit{Soundness}) The candidate models found by the sample-based search process can generate all the given traces, with the minimum number of changes to the incomplete model $\widetilde{M}$.

\textbf{Proof:} This is pretty straightforward as while generating the action set $B$ in the sample-based search method, the process checks to see if the conditions ($C_1(M)$, $C_2(M)$ and $C_3(M)$) are satisfied. It accepts the model as a candidate model only if these are satisfied.
%If these conditions fail then $B$ is generated such that models in the next steps can satisfy the traces.
Using Theorem 1, it can be said that if the sample-based search process finds a candidate model, the candidate model will be able to generate all the given traces. Uniform cost search ensures that the changes between the candidate models and the given incomplete domain model $\widetilde{M}$ are minimum.
Hence Proved.

\textbf{Theorem 4:} \textit{(Completeness)}
The sample-based search finds all the models satisfying the model test with the minimum number of changes to the incomplete model $\widetilde{M}$.

% Given sufficient traces such that they cover every operator, the heuristic based search will find all the models satisfying goal test with minimum changes (heuristic is complete).

\textbf{Proof:} 
We prove this by induction. Let us start from the first iteration through all the traces. In this iteration, since the unknown features are completely missing from the incomplete domain model $\widetilde{M}$, it makes the model a less constrained model than the complete model $M$ for achieving any goal. Hence, a trace will always be achieving the goal in $\widetilde{M}$ when it works under the complete model. This means $C_1(M)$ will always be satisfied. For any trace, $C_2(M)$ can still be false and in that case, it means that there is some action $a_i$ that is not well-justified under $\widetilde{M}$. To make that action well-justified, it is necessary that $a_i$ has some add effect that is a precondition for one of the subsequent actions in the trace. All these possibilities are included in the set $B$ discussed above.
\footnote{Note that in $B$, these possibilities are organized according to each trace and a new model is created for addressing each trace for the next iteration.} %corresponding to the failure of each trace.
% It can be seen that any other changes except these won't make the model satisfy this condition. 
These changes are necessary to ensure that model $M$ satisfies the conditions. 

Similarly, for any trace, $C_3(M)$ can also be false and in that case, the model $\widetilde{M}$ must have an optimal plan with shorter (lesser cost) than the trace. In that case, to satisfy $C_3(M)$, this optimal plan must be inexecutable in the complete model. This is necessarily achieved by updating $\widetilde{M}$ to add new preconditions to actions following the action $a_i$ up to which the trace and this optimal plan are the same. 
Here also, $B$ contains all the possibilities.
%in which predicate could be added to make the model satisfy the condition for each trace. 
Since a delete effect may also appear as a precondition in the same action, each option in $B$ that includes the addition of a new predicate to the preconditions comes with a paired option that includes the addition of the new predicate to the delete effects as well. 
%This includes all the possibilities where a predicate in the delete effect could be added.

Now let us assume that for the first $k$ iterations, all necessary changes to $\widetilde{M}$ with all possibilities are considered. For the $(k+1)^{th}$ iteration, since in the previous iterations we have added some new preconditions, the current (intermediate) model $M$ may no longer be less constrained than the complete model for achieving a goal. Hence, for any trace, it might not be achieving its goal in $M$, which means that $C_1(M)$ may fail. In this case, it is necessary that there is some action $a_i$ for which one of the preconditions is not satisfied. In this case, to make the trace executable, the only possibilities are to add the predicate (for the unsatisfied precondition) either to the add affects of some action before $a_i$ in the trace, or to the start state. Again we can see that $B$ includes all the possibilities in which predicate could be added to make the model satisfy this condition for each trace. For the $(k+1)^{th}$ iteration we can use similar reasoning as in the first iteration to argue that $B$ includes all the possibilities for $C_2(M)$ and $C_3(M)$ as well.

Hence, by induction, we show that the sample-based search includes all the possible ways in which the given incomplete model can be modified to satisfy all the given traces. The Uniform Cost Search (UCS) ensures that the changes are minimal. Hence, the sample-based search process will find all the candidate models with the minimum number of changes.

\subsection{Generating candidate models using online search}

In the real world, it is desirable to have an online search method that takes traces into account as they arrive. 
%Another way to perform the sample-based search is the online way where the models are searched using one trace at a time.
The search procedure is similar to the  sample-based method above and starts with the incomplete model $\widetilde{M}$.
%which is passed through the model test. 
The difference being that the model test is performed against just one trace at a time.
%corresponding to the current iteration. 
Within each iteration, the search continues till it finds $\mathcal{M}$. Once $\mathcal{M}$ is generated, it begins the next iteration by adding models in $\mathcal{M}$ (from the previous iteration) to the priority queue and then starting the search process again. It continues iterating until all available traces are checked. 
%The set of models $\mathcal{M}$ generated after the last iteration becomes the final set of candidate models.

In terms of computation, on average, the online sample-based search is expected to perform better. This is because in each iteration it uses only those models that satisfy the traces in the previous iterations. This restricts the number of models to be checked in each iteration since it is only constrained by one trace at any time. %The worst-case performance of the online search is the same as the previous the normal search method. The worst case is when the first trace is sufficient to learn the complete model. In that case, the online search method becomes equivalent to the normal search method.
It makes the online method highly dependent on the sequence of traces and not guaranteed to find the complete model. This is because it could find some incorrect model $M_1$, higher in the search-tree and at the same level reject some model $M_2$ that would have lead to the correct model by introducing a few more changes (deeper in the tree). In that iteration, the online search would not consider models that are deeper than $M_1$. Since in the next iteration only $M_1$ will be considered, there is a possibility that the correct model would never be found. 

\subsection{Planning under Incomplete Domain Knowledge}

After generating the set of candidate models $\mathcal{M}$, we find a robust plan for  $\widetilde{P}$ $=$ $\langle \widetilde{s_0}, g, \widetilde{M}, \widetilde{\zeta} \rangle$ such that it has the highest probability of achieving the goal under the weighted set of candidate models $\mathcal{M}$. Similar to \cite{Tuan}, we compile the problem of generating a robust plan into a Conformant Probabilistic Problem (CPP). A Conformant Probabilistic Problem \cite{PFF} is defined as $P’ = \langle I$, $g$, $D$, $\rho \rangle$ where $I$ is the belief over the initial state, $g$ is set of propositions that needs to be \textbf{T} for goal state, $D$ is the domain model and $\rho$ is the acceptable goal satisfaction probability. The domain model $D = \langle F'$, $A’ \rangle$, where $F’$ is the set of prepositions and $A’$ is the set of actions. Each $a’ \in A’$ has the set of preconditions $Pre(a’) \subseteq F’$ and $E(a’)$, the set of conditional effects. Each $e’ \in E(a’)$ is a pair of $con(e’)$ and \.{o}$(e’)$ where $con(e’) \subseteq F’$ which enables $e’$ and \.{o}$(e’)$ is a set of outcomes $\epsilon$. The outcome $\epsilon$ is a triplet $\langle Pr(\epsilon)$, $add(\epsilon)$, $del(\epsilon) \rangle$ where $add(\epsilon)$ adds prepositions to and $del(\epsilon)$ deletes prepositions from the current model with probability $Pr(\epsilon)$.

% $del(\epsilon)$ $\subseteq$ $F’$/deletes prepositions to/from the current state with probability $Pr(\epsilon)$.

% We use Probabilistic-FF which is a state-of-the-art Conformant Probabilistic Planner[10] for generating plan for CPP. 

A compilation that translates the original planning problem $\widetilde{P}$ to a conformant probabilistic planning problem $P’$ is defined as follows:

\begin{itemize}
\item

For each candidate model $M_i \in \mathcal{M}$ a preposition $m_i$ is introduced. Let the set of these prepositions be $\widehat{M}$. Further, a set $\widehat{F} = \bigcup\limits_{i=1}^n F_i$ is introduced, where $F_i$ is the set of prepositions instantiated by predicates $R_i \in M_i$ where $M_i \in \mathcal{M}$ and n is the total number of model in $\mathcal{M}$.
For the compiled problem set of prepositions $F' = \widehat{M} \cup \widehat{F}$. 

\item
For each model $M_i \in \mathcal{M}$, $U'_i$ is created which is the set of new prepositions that were not present in $\widetilde{M}$. Using this, now a domain model is created $D$ from $\widetilde{M}$ as follows:

\begin{itemize}
\item A new action $a'_0$ that initializes the start state with new/missing prepositions is introduced. The action $a'_0$ is defined as $Pre(a'_0) = \emptyset$. $\forall m_i \in \widehat{M}$, a conditional effect $e'_i \in E(a'_0)$ is created such that $con(e'_i) = \{m_i\}$ and each outcome $\epsilon_j \in$ \.{o}$(e'_i)$ has $add(\epsilon_j) = \mu'$ and $del(\epsilon_j) = \emptyset$ where $\mu' \in {\mathfrak P} (U'_i)$. For each outcome, $Pr(\epsilon_j) = 1/|{\mathfrak P} (U'_i)|$. This essentially initializes the start state for each model $M_i$, considering all the possibilities for new prepositions to be present in it with equal probability. In the packing domain, if $U'_i$ $=$ $\{$ {\tt(pred\_0 i1)}, {\tt(pred\_0 i2)}$\}$, all the possibilities are $\{\}$, $\{${\tt(pred\_0 i1)}$\}$, $\{${\tt(pred\_0 i1)}$\}$, $\{${\tt(pred\_0 i1)}, {\tt(pred\_0 i2)}$\}$. All of these will be considered to be present in the start state with equal probability for each, which is $0.25$ here.

\item For each action $a \in \widetilde{A}$ in $\widetilde{M}$, if model $M_i \in \mathcal{M}$ adds a preposition $u'_i \in U'_i$ to $\widetilde{Pre(a)}$, a conditional rule $e_i \in E(a)$ is created, such that $con(e_i) = \widetilde{Pre(a)} \cup \{m_i\} \cup \{u'_i\}$, $add(e_i) = \widetilde{Add(a)}$ and $del(e_i) = \widetilde{Del(a)}$, $m_i \in \widehat{M}$ is the binary predicate corresponding to $M_i$. For example, if in model $M_i$, action {\tt place} {\tt i1} has the new preposition {\tt(pred\_0} {\tt i1)} in it's precondition, then the action in the compiled domain will have a conditional rule where $con(e_i) = \widetilde{Pre({\tt place}\,{\tt i1})} \cup \{m_i\} \cup \{${\tt(pred\_0} {\tt i1)}$\}$ and $add(e_i)$ and $del(e_i)$ will remain the same.

\item For each action $a \in \widetilde{A}$ in $\widetilde{M}$, if model $M_i \in \mathcal{M}$ adds a preposition $u'_i \in U'_i$ to $\widetilde{Add(a)}$, a conditional rule $e_i \in E(a)$ is created, such that $con(e_i) = \widetilde{Pre(a)} \cup \{m_i\}$, $add(e_i) = \widetilde{Add(a)} \cup \{u'_i\}$, $del(e_i) = \widetilde{Del(a)}$, $m_i \in \widehat{M}$, is the binary predicate corresponding to $M_i$.

\item For each action $a \in \widetilde{A}$ in $\widetilde{M}$, if model $M_i \in \mathcal{M}$ adds a preposition $u'_i \in U'_i$ to $\widetilde{Del(a)}$, a conditional rule $e_i \in E(a)$ is created, such that $con(e_i) = \widetilde{Pre(a)} \cup \{m_i\}$, $add(e_i) = \widetilde{Add(a)}$, $del(e_i) = \widetilde{Del(a)} \cup \{u'_i\}$, $m_i \in \widehat{M}$, is the binary predicate corresponding to $M_i$.
\end{itemize}

The modified domain becomes $D$ for the problem $P’$. 

\item
The initial belief state $I = (\bigwedge\limits_{f \in \widetilde{s_0}} f) \land ( oneof (m_1, m_2, ... m_n))$ where $oneof$ returns \textbf{T} when exactly one of its input is \textbf{T}. The probability of each $m_i$ is equal to its weight in $\mathcal{M}$ and all the other prepositions are certain. 

\item
In the compiled problem, $\rho'$ = $\rho$ represents the probability of success of the conformant plan generated in the problem $\widetilde{P}$.
\end{itemize}

The process of generating the robust plan is quite straightforward. Initially, $\rho$ = 1, and a conformant plan is calculated for the given problem using conformant probabilistic planner. If a conformant plan is not found $\rho$ is decreased by $\Delta$ until one is found. The plan so obtained is a robust plan for problem $\widetilde{P}$ $=$ $\langle \widetilde{s_0},$ $g,$ $\widetilde{M}, \widetilde{\zeta} \rangle$ and has the highest probability of success given the weighted set of candidate models $\mathcal{M}$.
%$\oplus$
% The application of action $a'_0 \in A'$ creates more models from each $M_j$ by adding new/missing prepositions in the start state. Let the set of all these new models be $\|M'\|$.  The goal satisfiability $\rho$ in our compiled problem represents the fraction of models in $\|M'\|$ that have a conformant plan for the problem $\widetilde{P}$. Initially $\rho$ is set to 1.0 and if conformant plan is not found it is decreased by $\Delta$ until one is found.

\textbf{Theorem 5:} If $\pi' = \langle a'_0, a'_1, a'_2, a'_3, ... a'_n \rangle$ is a plan for the complied problem $P'$ with goal probability $\rho$ then, $\rho$ is also the probability of success of the plan $\pi = \langle a_1, a_2, a_3, ... a_n \rangle$ in the problem $\widetilde{P}$.
% the fraction of models in $\mathcal{M'}$ for which $\pi = \langle a_1, a_2, a_3, ... a_n \rangle$ is a valid plan for problem $\widetilde{P}$ is $\rho$.

\textbf{Proof:} The compilation defines a bijective mapping between each state $i_{j} \in I$ of the problem $P'$ and each model $M_j \in \mathcal{M}$ of the problem $\widetilde{P}$. Also, the probability of $i_{j}$ in $I$ is same as the probability of $M_j$ in $\mathcal{M}$.
Let the belief state after execution of action $a'_0$ in $I$ be $b_{s'_{0}}$. 
The action $a'_0$ initializes multiple start states for each model in $\mathcal{M}$. One can think of it as generating multiple sub-models for each $M_j \in \mathcal{M}$ based on different start states. Let the set of all these sub-models be $\mathcal{M'}$.
It is easy to see that there is a bijective mapping between each $s'_{0j} \in b_{s'_{0}}$ and model $M'_j \in \mathcal{M'}$ with same probability as well. Moreover, if $s_{0j}$ is the start state for model $M'_j \in \mathcal{M'}$, any prepositions $p \in s'_{0j}$ iff $p \in s_{0j}$. %Ignoring action $a'_0$,
The application of plan $\pi' \setminus \{a'_0\}$ in belief state $b_{s'_{0}}$ generates sequence of belief states $\langle b_{s'_{1}}, b_{s'_{2}}, ... b_{s'_{n}} \rangle$. Similarly, executing $\pi$ in $s_{0j}$ for model $M'_j$ generates state sequence $\langle s_{1j}, s_{2j}, ... s_{nj} \rangle$. 
To any state $s'_{ij} \in b_{s'_{i}}$, every action $a' \in A'$ adds/deletes same set of prepositions against the same conditions as action $a \in A$ to $s_{ij}$ for $M'_j \in \mathcal{M'}$. Hence, using induction one can say that for any state in the sequences mentioned above preposition $p \in s'_{kj}$ iff $p \in s_{kj}$ $(\forall k \in \{1, 2, ... n\})$. Therefore, $s_{nj} \supseteq g$ iff $s'_{nj} \supseteq g$. 
Since the all actions (except the dummy action $a'_0$) are deterministic, if plan $\pi'$ achieves goal for $s'_{0j} \in b_{s'_{0}}$, then $\pi$ achieves goal for $s_{0j}$ in $M'_j \in \mathcal{M'}$. Hence, if $\pi'$ achieves goal with probability $\rho$ in $P'$, then the probability of success of $\pi$ in the problem $\widetilde{P}$ is $\rho$. Hence Proved. 

\begin{table*}[]
\resizebox{\textwidth}{!}{%
\begin{tabular}{|l|c|r|r|r|c|c|c|r|r|r|c|c|c|l|l|}
\hline
\multirow{2}{*}{Doms} & \multirow{2}{*}{\begin{tabular}[c]{@{}l@{}}\# T\end{tabular}} & \multicolumn{3}{c|}{Model Searched} & \multicolumn{3}{c|}{Candidate Models} & \multicolumn{3}{c|}{Time(secs)} & \multicolumn{3}{c|}{Plan Success} & \multicolumn{2}{c|}{\begin{tabular}[c]{@{}l@{}}Avg Cost Inc\end{tabular}} \\ \cline{3-16} 
 &  & BF & SS & OS & BF & SS & OS & BF & SS & OS & SS & OS & Default & SS & OS \\ \hline
\multicolumn{16}{|l|}{One predicate missing at a time} \\ \hline
\multirow{3}{*}{Rovers} & 3 & 1500 & 400 & 400 & 2 & 2 & 2 & 205.78 & 23.32 & 37.80 & 8/8 & 8/8 & 0/8 & +0.25 & +0.25 \\ \cline{2-16} 
 & 3 & 1500 & 200 & 200 & 2 & 2 & 2 & 47.62 & 10.56 & 10.98 & 8/8 & 8/8 & 0/8 & +0.63 & +0.63 \\ \cline{2-16} 
 & 5 & 10300 & 1400 & 300 & 3 & 3 & 3 & 520.29 & 103.89 & 184.52 & 11/11 & 11/11 & 0/11 & +0.64 & +0.64 \\ \hline
\multirow{3}{*}{Miner} & 2 & 700 & 30 & 30 & 1 & 1 & 1 & 8.74 & 0.90 & 1.12 & 12/12 & 12/12 & 0/12 & +1.25 & +1.25 \\ \cline{2-16} 
 & 4 & 2520 & 430 & 70 & 1 & 1 & 1 & 517.69 & 156.14 & 31.07 & 10/10 & 10/10 & 0/10 & +1.00 & +1.00 \\ \cline{2-16} 
 & 2 & 140 & 10 & 10 & 1 & 1 & 1 & 2.42 & 0.49 & 0.62 & 12/12 & 12/12 & 0/12 & +1.25 & +1.25 \\ \hline
\multicolumn{16}{|l|}{Two predicates missing at a time} \\ \hline
\multirow{3}{*}{Rovers} & 3 & - & 26500 & 26600 & - & 4 & 4 & - & 645.11 & 484.96 & 7/7 & 7/7 & 0/7 & +0.57 & +0.57 \\ \cline{2-16} 
 & 6 & - & 297000 & 31900 & - & 3 & 3 & - & 8873.6 & 1484.72 & 8/8 & 8/8 & 0/8 & +0.25 & +0.25 \\ \cline{2-16} 
 & 6 & - & 250700 & 32000 & - & 3 & 3 & - & 8456.40 & 1671.37 & 8/8 & 8/8 & 0/8 & +0.00 & +0.00 \\ \hline
\multirow{3}{*}{Miner} & 4 & - & 21900 & 3600 & - & 1 & 1 & - & 1023.41 & 192.62 & 10/10 & 10/10 & 0/10 & +1.00 & +1.00 \\ \cline{2-16} 
 & 4 & - & 6600 & 620 & - & 1 & 1 & - & 450.05 & 41.72 & 10/10 & 10/10 & 0/10 & +1.00 & +1.00 \\ \cline{2-16} 
 & 2 & 121300 & 260 & 260 & 1 & 1 & 1 & 1610.92 & 3.29 & 2.83 & 12/12 & 12/12 & 0/12 & +1.25 & +1.25 \\ \hline
\multicolumn{16}{|l|}{Three predicates missing at a time} \\ \hline
Rovers & 6 & - & - & 1175000 & - & - & 3 & - & - & 25610.00 & - & 8/8 & 0/8 & - & +0.25 \\ \hline
Miner & 4 & - & 451500 & 11420 & - & 1 & 1 & - & 11104.00 & 364.22 & 10/10 & 10/10 & 0/10 & +1.00 & +1.00 \\ \hline
\end{tabular}}
\caption{Results from synthetic domains }
\label{table:1}
\end{table*}

\section{EVALUATION}
We performed two types of experiments. In the first experiment, we tested our algorithm on various International Planning Competition (IPC) domains. In the second experiment, we created a more complex version of our packing domain to show the practical benefits of our algorithm. Plans were generated with Fast-Downward planner \cite{FDplan}, and for solving conformant probabilistic planning problems, probabilistic-FF planner \cite{PFF} was used. 

% The evaluations were done using two types of experiments. In the first one we have tested our algorithm on various International Planning Competition (IPC) domains. In the second type of experiment we simulated a complex version of our packing domain to test our algorithm. To generate optimal plans we have used Fast-Downward planner \cite{FDplan} and for solving Conformant Probabilistic Planning problem we have used Probabilistic-FF planner \cite{PFF}. 

% We noticed that if the missing predicate is present in the init state ($s_0$), the number of models in the model search as well as the number of candidate models increases tremendously. This not only increases the search time but also make it almost impossible of the basic version of PFF to generate an output. Hence in out tests we assume that the missing predicate is not present in the init state. Note the our solution will still work if the missing predicate is present in the start state since the formulation considers it.

\subsection{Synthetic Domains}

For synthetic evaluation, we have used two domains. In the \textit{rover} domain, there are multiple rovers each equipped with  capabilities like sampling soil, rock and capturing images to be used at different waypoints. The second domain is a slightly modified version of the \textit{gold-miner} domain. In this domain, we have a grid and the task is to pick up gold from a particular cell in the grid and deposit it in another. Some cells have a laser or a bomb that could be used to clear the blocked cells. We introduced incompleteness by deleting some predicates that could be generated by one action and are preconditions to some other actions. For example, in rover domain, the precondition to capture an image is that the camera should be calibrated.
For each domain, we created a few incomplete domains and problems by deleting either a single or multiple (up to 3) predicates. For this experiment, we have deleted only those predicates that were not present in the initial state. Using the complete domain, optimal plans were generated which were used as the teacher traces. These traces were then projected onto the incomplete model and were given to the agent. %to the algorithm.
%The robust plan generated by the algorithm was verified by checking its execution against the complete domain.

Table \ref{table:1} shows the detailed results of our experiments. For each incomplete domain, the results are presented for all the $3$ methods: Brute Force (BF), Sample-based Search (SS) and Online sample-based Search (OS). Here each incomplete domain was given sufficient traces such that the robust plan generated in the end was successful for every test case. This happens when we have very few candidate models. 
%For OS, multiple random sequences of traces were considered and the one with the least time is shown in the table. 
\say{-} in the table represents the situation where the model search time exceeded the predefined limit for that domain. Since the brute-force approach is very expensive, it timed-out in almost every domain tested when multiple predicates were missing. It can clearly be seen that sample-based search (either SS or OS) reduces the number of models to be searched by a considerable amount. We can also see that as the number of missing predicates increases, the time taken by the algorithm to generate candidate models increases by a significant amount which is due to the exponential increase in the search space of the possible models. In such cases, we can see that the online search method performed much better than the sample-based search method except for cases where only $1$ predicate was missing. We can also see that without using our method and running the planner on the incomplete domain, the plan almost always fails. It can also be seen that the plan generated by our algorithm incurs a little more cost than the optimal plan in the complete domain. 

\begin{figure}[h]
    \centering
    \includegraphics[width=8cm]{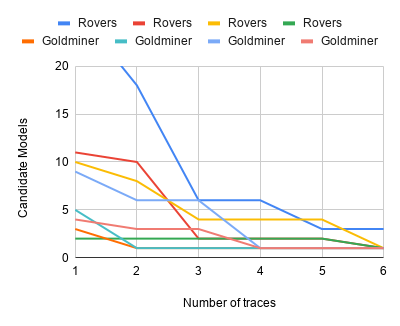}
    \caption{Candidate models VS number of Traces}
    \label{fig:graph}
\end{figure}

Fig. \ref{fig:graph} shows the variation of the number of candidate models generated with the number of traces. It can clearly be seen that as the number of traces increase, we get fewer but more accurate candidate models.

\begin{figure*}[h]
    \centering
    \includegraphics[width=17.6cm]{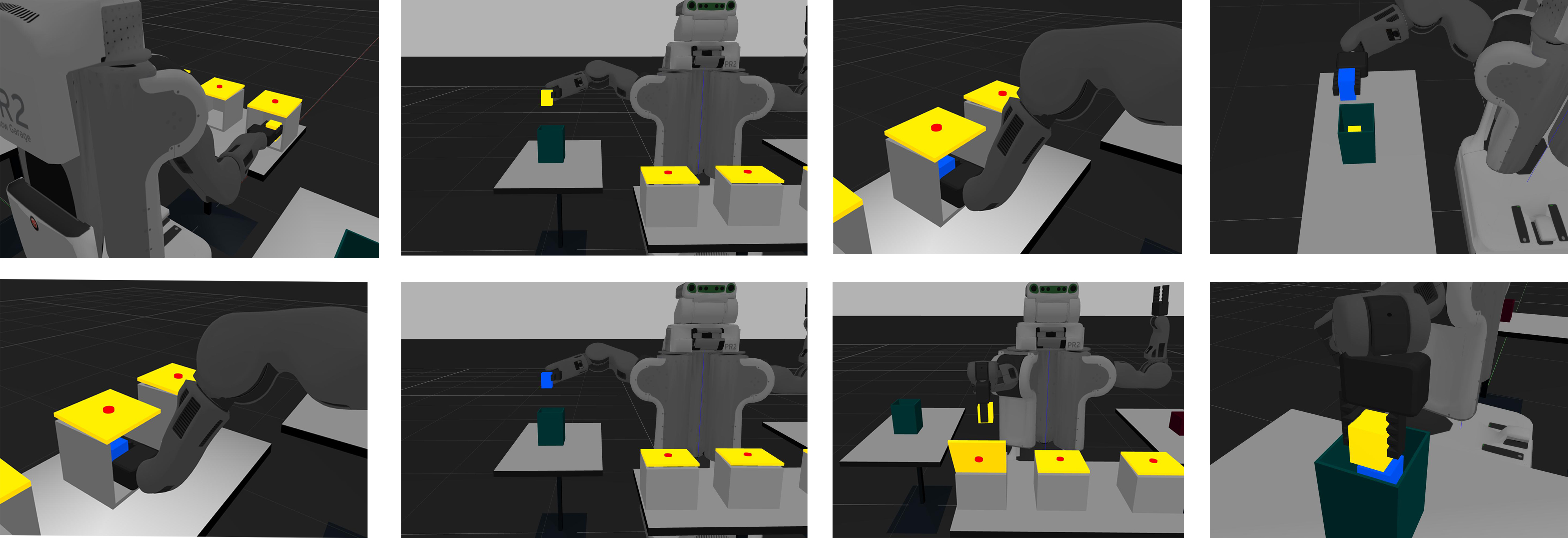}
    \caption{ Comparison of action sequence generated with a standard planning method (top) and our method (bottom). 
    }
    \label{fig:simulation}
\end{figure*}

\subsection{Simulated Domain}
In this experiment, we have created a simulated robotics domain which is a more complex version of our motivating packing domain. In this domain instead of one now we have two constraints for putting items in boxes. As before, the first constraint is that a fragile item cannot be stacked. The second constraint is that a fragile item cannot be dropped into the box and instead it has to be placed carefully. Here, we have two grasping actions: {\tt horizontal\_grasp} and {\tt vertical\_grasp}. For {\tt horizontal\_grasp}, the surroundings of the item to be picked up should be clear. For {\tt vertical\_grasp}, clear surroundings are not a necessity.
But in situations where the item to be picked is placed in the container (as shown in Fig. \ref{fig:simulation}), the container needs to be opened first by pressing a button on the top. On the other hand, this is not needed for {\tt horizontal\_grasp}. Furthermore, if the robot has picked up an item horizontally, it is constrained to use {\tt drop} action instead of {\tt place}. For vertical grasp, both {\tt place} and {\tt drop} are possible. The goal of the robot is the same as before. %to put all the items into the boxes by using the minimum number of boxes.

% The incompleteness in this domain is due to the missing of {\tt not\_fragile} predicate, which specifies that the given item is not fragile. Hence, neither the robot has any knowledge about the presence of this predicate nor it has the ability to sense that. 
% Notice that in this case, the missing predicate will also be present in the start state since specific types of items are always fragile.

% We generated optimal traces using the complete domain model. The algorithm used these traces along with the incomplete domain model to find candidate models and then generated a robust plan. The robust plan generated was verified by executing it in the complete domain model.  
% {\color{red}{explain the results: the steps the robot takes with different algorithms.}}

Fig. \ref{fig:simulation} shows the setup of our simulated experiments. The left sequence (top to bottom) is the one that did not use our algorithm. It can be seen that the robot was not able to distinguish between fragile items (in yellow) and non-fragile items (in blue). Hence, it used {\tt horizontal\_grasp} for picking up the fragile item and {\tt drop} to put the fragile item into the box, which could damage the fragile item. Furthermore, in the subsequent actions, the robot stacked an item over the fragile item, which was also undesirable. On the other hand, in the right sequence showing the actions executed using our algorithms (SS and OS), the robot first picked up the non-fragile item using {\tt horizontal\_grasp} and then put it into the box using the {\tt drop} action. Then it used {\tt vertical\_grasp} followed by {\tt place} to stack the fragile item carefully over non-fragile item. This was a successful sequence as none of the constraints of the complete domain were violated.   
%We tested our algorithms on a few more scenarios where using our algorithm.
This experiment showed that the robot was able to learn what types of items were not fragile and acted accordingly,
with such knowledge completely missing in the domain initially.

% \subsection{Results and Analysis}

\section{CONCLUSIONS}

In this paper, we have formally introduced the problem of \textit{Domain Concretization} and have discussed its prevalence. We have presented a solution that uses teacher traces and the incomplete domain model to generate a set of candidate models and then finds a robust plan that achieves the goal under the maximum number of  candidate models. We have formulated the model search process and developed a sample-based search to make the search more efficient. For practical use, we have also presented an online version of this search method where we used one trace at a time to refine our candidate models. Our methods were tested on IPC domains and a simulated robotics domain where our methods significantly increased plan success rate.

% This problem opens up several research directions and few of them are mentioned here. The process of candidate model generation, even with sample-based and online methods, is quite computationally expensive, especially when the number of missing predicates increases. Finding more efficient heuristics and constraints would be an interesting future direction to work on. Another possible future direction could be to consider solving the problem in a better iterative process something similar to the online method. In the online method presented here, if an intermediate step learns only incorrect models and not any correct model then the next steps fail to learn the correct model. Future work could include solving this problem and enabling the algorithm to learn the correct model even if some previous steps learn only incorrect domain models.

% The current work is limited to deterministic domains. Solving the problem of domain concretization for stochastic domains could be another possible future direction. Furthermore, currently, we assume that the human model is complete. Considering the incomplete human model with a complete robot's model where the robot's model could be used to refine the human model could be a very interesting future direction to work on. This could be further extended to the interactive scenarios where the human and the robot works as a team, both having incomplete domain models.

\addtolength{\textheight}{-12cm}   % This command serves to balance the column lengths
                                  % on the last page of the document manually. It shortens
                                  % the textheight of the last page by a suitable amount.
                                  % This command does not take effect until the next page
                                  % so it should come on the page before the last. Make
                                  % sure that you do not shorten the textheight too much.

%%%%%%%%%%%%%%%%%%%%%%%%%%%%%%%%%%%%%%%%%%%%%%%%%%%%%%%%%%%%%%%%%%%%%%%%%%%%%%%%

%%%%%%%%%%%%%%%%%%%%%%%%%%%%%%%%%%%%%%%%%%%%%%%%%%%%%%%%%%%%%%%%%%%%%%%%%%%%%%%%

%%%%%%%%%%%%%%%%%%%%%%%%%%%%%%%%%%%%%%%%%%%%%%%%%%%%%%%%%%%%%%%%%%%%%%%%%%%%%%%%
%\section*{APPENDIX}

%\section*{ACKNOWLEDGMENT}

\bibliographystyle{IEEEtran}
\bibliography{dcfe}

\end{document}